\documentclass[letterpaper, 10 pt, conference]{ieeeconf}

\IEEEoverridecommandlockouts    

\usepackage{times} 
\usepackage{graphicx} 
\usepackage[ruled]{algorithm2e}  
\usepackage[utf8]{inputenc} 
\usepackage{setspace} 
\usepackage{hyperref} 
\usepackage[noadjust]{cite}
\usepackage{flushend} 

\usepackage{amsmath,amsfonts,amssymb,amsopn,bm}

\renewcommand{\P}{\mathbb{P}} 
\newcommand{\1}{\mathbf{1}}
\newcommand{\E}{\mathbb{E}}
\newcommand{\R}{\mathbb{R}}

\newcommand{\twonormsq}[1]{\|#1\|_2^2}

\newcommand{\xxnote}[3]{}
\ifx\hidenotes\undefined
  \usepackage{color}
  \renewcommand{\xxnote}[3]{\color{#2}{#1: #3}}
\fi

\renewcommand{\baselinestretch}{0.975}

\newcommand\figref{Fig.~\ref}

\usepackage{paralist}
\title{\LARGE \bf
Leveraging Post Hoc Context for Faster Learning in Bandit Settings \\
with Applications in Robot-Assisted Feeding
}

\author{Ethan K. Gordon$^{1}$,
Sumegh Roychowdhury$^{2}$,
Tapomayukh Bhattacharjee$^{1}$, \\
Kevin Jamieson$^{1}$, and
Siddhartha S. Srinivasa$^{1}$
\thanks{$^{1}$ Ethan K. Gordon, Tapomayukh Bhattacharjee, Kevin Jamieson, and Siddhartha S. Srinivasa are with the Department of Computer Science and Engineering, University of Washington, Seattle, WA 98195
        {\tt\small \{ekgordon, tapo, jamieson, siddh\}@cs.washington.edu}}%
\thanks{$^{2}$Sumegh Roychowdhury is with the Indian Institute of Technology Kharagpur, Kharagpur, India, 
        {\tt\small sumegh01@iitkgp.ac.in}. Work done as a UW intern.}%
}

\begin{document}

\maketitle
\thispagestyle{empty}
\pagestyle{empty}

\begin{abstract}
Autonomous robot-assisted feeding requires the ability to acquire a wide variety of food items. However, it is impossible for such a system to be trained on all types of food in existence. Therefore, a key challenge is choosing a manipulation strategy for a previously unseen food item. Previous work showed that the problem can be represented as a linear bandit with visual context. However, food has a wide variety of multi-modal properties relevant to manipulation that can be hard to distinguish visually. Our key insight is that we can leverage the haptic context we collect during and after manipulation (i.e., ``post hoc'') to learn some of these properties and more quickly adapt our visual model to previously unseen food. In general, we propose a modified linear contextual bandit framework augmented with post hoc context observed after action selection to empirically increase learning speed and reduce cumulative regret. Experiments on synthetic data demonstrate that this effect is more pronounced when the dimensionality of the context is large relative to the post hoc context or when the post hoc context model is particularly easy to learn. Finally, we apply this framework to the bite acquisition problem and demonstrate the acquisition of 8 previously unseen types of food with 21\% fewer failures across 64 attempts.
\end{abstract}

\section{INTRODUCTION}
Many of us take eating for granted, but approximately 1.0 million people in the US alone cannot eat without assistance \cite{2012Brault}. Autonomous robot-assisted feeding could save time for caregivers and increase people's sense of self worth \cite{1990Prior, 1994Stanger}. Existing robotic feeding systems \cite{myspoon, obi} rely on preprogrammed movements, making it difficult for them to adapt to environmental changes. In general, a robust feeding system must be able to acquire a bite of food in an uncertain environment, a task known as ``bite acquisition.'' This work focuses on the acquisition of food items that the robot may not have seen or manipulated before.

Previous work has identified useful manipulation strategies for a variety of food items \cite{bhattacharjee2018food} and ways to generalize those strategies to similar-looking food items \cite{2019Feng, 2019Gallenberger}, but one challenge is figuring out how to select a manipulation strategy for previously unseen food. Recent work has suggested modeling bite acquisition as a contextual bandit \cite{gordon2020adaptive}. The robot can observe visual context for each food item, select from a set of discrete manipulation strategies, and observe partial (or bandit) feedback in the form of a binary success or failure.

\begin{figure}[t!]
    \centering
    \includegraphics[width=\linewidth]{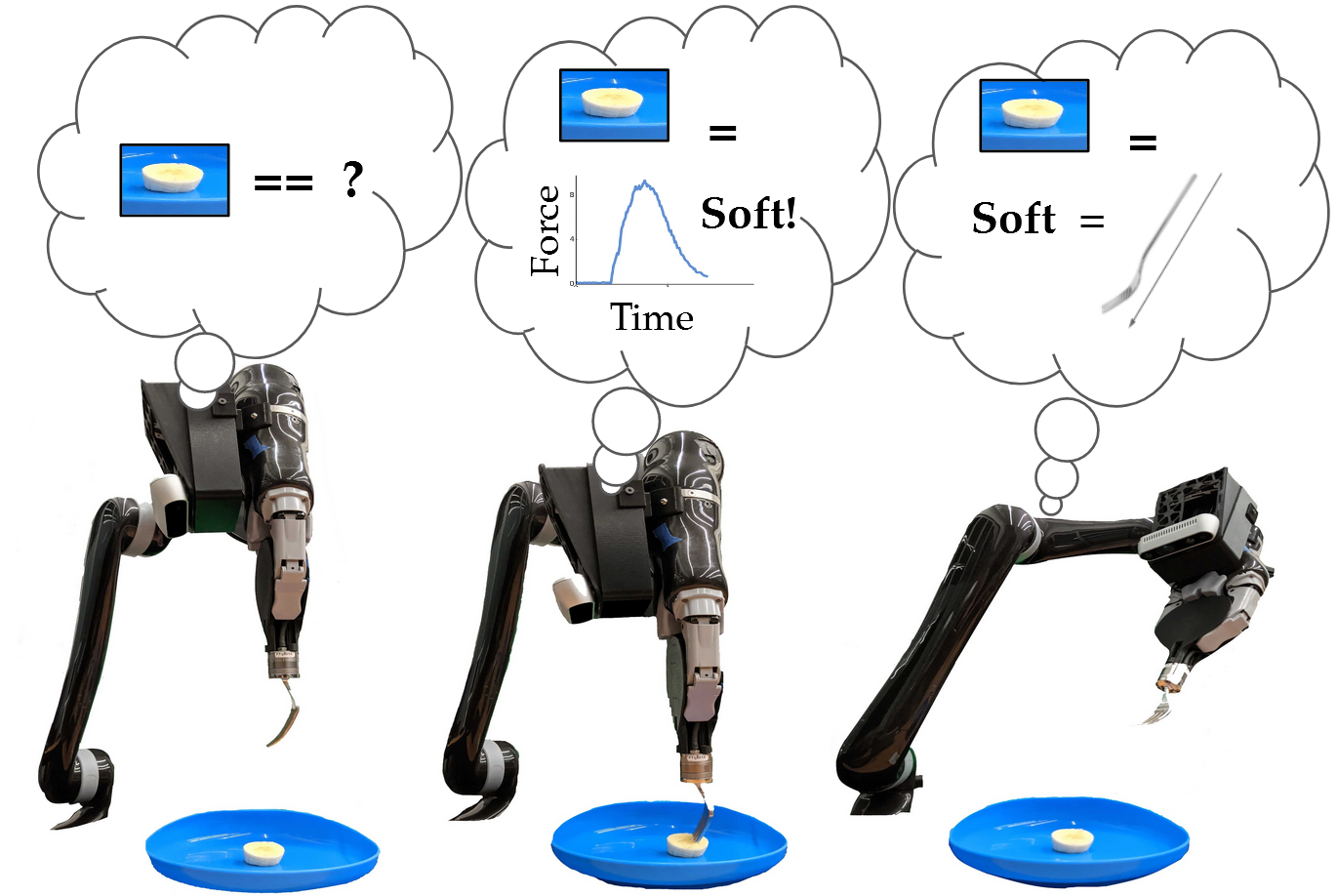}
    \vspace{-0.6cm}
    \caption{The Assistive Dextrous Arm (ADA) can leverage post hoc context to quickly figure out which action is best for picking up a banana slice. \emph{Left:} Banana slices have never been seen before. Acquired visual context. \emph{Middle:} Haptic data suggests that banana slices are soft. \emph{Right:} Combining visual and haptic contexts, appropriate action is predicted.
    }
    \label{fig:intro}
    \vspace{-0.7cm}
\end{figure}
\begin{figure*}[t!]
    \centering
    \includegraphics[width=0.95\linewidth]{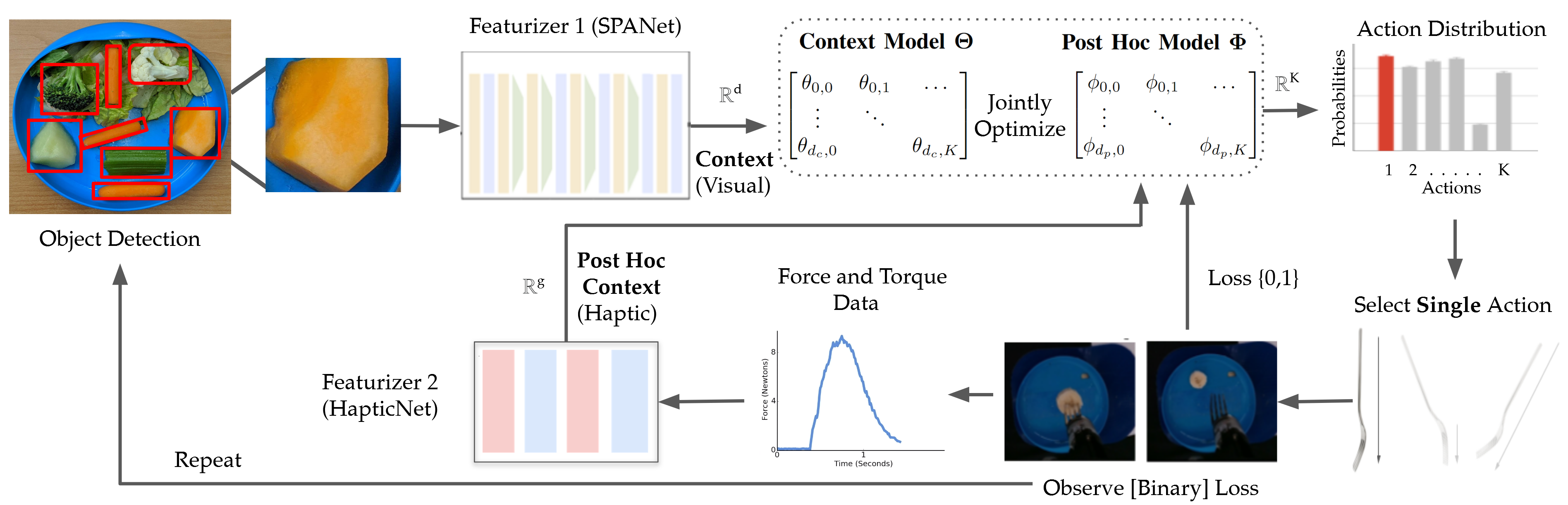}
    \vspace{-0.3cm}
    \caption{Post hoc augmented contextual bandit framework. We only observe the visual context from SPANet prior to action selection, but the post hoc context from HapticNet is used with the observed loss to update the visual model.}
    \label{fig:system}
    \vspace{-0.6cm}
\end{figure*}
However, the method presented in that work required about 10 failures over 25 attempts to learn the optimal manipulation strategy for a given new food item. Visual context is only a single mode. Food items that look similar (but not identical), such as ripe and un-ripe banana slices, can have very different consistencies, leading to different optimal manipulation strategies. It is hard to learn this map with only visual data. Haptic feedback from physical interactions can be informative for object classification~\cite{SchneiderSturmStachniss2009, AllenRoberts1989, bhattacharjee2018food} and inferring object properties such as haptic adjectives~\cite{chu2015robotic}, rigidity~\cite{DrimusKootstraBilbergKragic2011}, elasticity~\cite{FrankSchmeddingStachnissTeschnerBurgard2010a}, hardness~\cite{TakamukuGomezHosodaPfeifer2007}, and compliance~\cite{kaboli2014humanoids, bhattacharjee2017inferring}. Previous work has also shown that combining visual and haptic modalities can help towards inferring global haptic mapping~\cite{shenoi2016crf} and learning multi-modal representations~\cite{lee2019making} during manipulation. Our key insight is that \emph{we can leverage haptic feedback collected after action selection during manipulation to more quickly learn how to map visual information to the optimal strategy for a given type of food}.

More generally, we propose augmenting the traditional contextual bandit framework with post hoc, multi-modal context collected after action selection. We show empirically that when the post hoc context is relatively simple to model, it can be used to more quickly model the regular context, leading to faster learning and lower cumulative regret. Our major contributions are (1) a proposed modification to LinUCB \cite{Chu2011} and online linear regression to handle post hoc context, (2) experimental results on synthetic data that provide insights on the efficacy of this framework, and (3) empirical evidence that demonstrates improvement over the traditional contextual bandit setting in a real robot bite acquisition task relevant to the application of robot-assisted feeding. Our current action space is small, consisting of only 6 discrete strategies, but future work can leverage these insights to examine a larger action space that can handle an even wider variety of food items and realistic plates.
\section{RELATED WORK}
\subsection{Online Learning and Contextual Bandits}
\label{sec:conban}
Contextual bandit algorithms have seen widespread success in health interventions \cite{Klasnja2015,Hochberg2016}, online advertising \cite{Tang2013,Bottou2013}, adaptive routing \cite{Awerbuch2004}, clinical trials \cite{Shortreed2011}, music recommendations \cite{Wang2014}, education \cite{Mandel2014}, and financial portfolio design \cite{Shen2015}. Adoption in robotics has included selecting trajectories for object rearrangement \cite{Koval2015}, kicking strategies in robotic soccer \cite{Mendoza2016}, selecting among deformable object models for acquisition tasks \cite{McConachie2017}, and the aforementioned work on selecting manipulation strategies for deformable food items \cite{gordon2020adaptive}. All of these applications leverage a single context vector observed prior to action selection and the scalar incurred loss or reward. We propose leveraging higher dimensional feedback observed after action selection to speed up learning.

Baseline exploration strategies include epoch-greedy \cite{Langford2008}, LinUCB \cite{Li2010}, RegCB \cite{Foster2018} and Online Cover \cite{Agarwal2014}. For a recent and thorough overview, we refer the interested reader to \cite{Lattimore2019,Bietti2018}. Our work is distinct from bandits with delayed feedback \cite{Vernade2018-jv} in that the post hoc context is not delayed by any time steps, but just observable after action selection. Our work could potentially be compared with the bandits-with-expert-advice setting and associated algorithms like EXP4 \cite{Auer2002a}, in the sense that the context model and the post hoc context model could be thought of as competing action recommendations, though the experts in this setting generally make their predictions exclusively prior to action selection.

\subsection{Robot-Assisted Feeding: Food Manipulation}
General food manipulation has been studied in various environments, such as the packaging industry~\cite{chua2003robotic, erzincanli1997meeting, morales2014soft, brett1991research, williams2001teaching, blanes2011technologies}. These tend to focus on the design of application-specific grippers for robust sorting and pick-and-place. Other work shows the need for visual sensing for quality control~\cite{brosnan2002inspection, du2006learning, ding1994shape} and haptic sensing for grasping deformable food items without damaging them~\cite{chua2003robotic, erzincanli1997meeting, morales2014soft, brett1991research, williams2001teaching, blanes2011technologies}. Research labs have also explored meal preparation~\cite{ma2011chinese, sugiura2010cooking}, baking cookies~\cite{bollini2011bakebot}, making pancakes~\cite{beetz2011robotic}, separating Oreos~\cite{oreovideo}, and preparing meals~\cite{gemici2014learning} with robots. Most of these studies either interacted with a specific food item with a fixed manipulation strategy~\cite{bollini2011bakebot, beetz2011robotic} or with an unchanging set of food items and manipulation strategies~\cite{2019Feng, 2019Gallenberger, 2016Park, herlant_thesis}. Some of these studies have looked at using multi-modal data~\cite{gemici2014learning} or online learning~\cite{gordon2020adaptive}, but not a combination of the two.

Our visual context is generated using the \emph{Skewering Position Action Network} (SPANet) from \cite{2019Feng} while our action space and our haptic context specification are derived from human data~\cite{bhattacharjee2018food}.

\section{PRELIMINARY: CONTEXTUAL BANDITS}
Previous work \cite{gordon2020adaptive} showed the utility of representing the bite acquisition setting as a contextual bandit. For each attempt, the agent observes $d_c$-dimensional visual context (an image of the food) and selects from a discrete set of $K$ manipulation strategies, observing a loss of 0 on success and 1 on failure. Here we cover the specifics of this setting that are relevant to our proposed augmentation.

\subsection{Formulation}
General online supervised learning has an agent learn a map $f : \R^{d_c} \rightarrow \R^K$ between a $d_c$-dimension context vector $c$ and a $K$-dimension loss vector $l$ given a sample $(c_t, l_t)$ at each time step $t$. In a discrete interactive learning setting, the agent will first observe the context $c_t$, choose an action $a_t \in [K]$, and then observe the full loss vector $l_t$ while incurring loss $l_t[a_t]$. The agent's goal in this setting is to minimize \emph{cumulative regret}
\begin{align}
R_T := \max_{a'}\sum_t^T \left(l_t[a_t] - l_t[a']\right)
\end{align}
the difference between the loss incurred by the agent and the lowest loss it was possible to incur.

In a contextual bandit setting, the agent is restricted to \emph{bandit feedback}: observing only the loss incurred $(l_t[a_t])$ rather than the full loss vector $l_t$. This creates a trade-off between exploring actions we are unsure about and exploiting actions likely to incur little loss. In general, a contextual bandit algorithm consists of two parts: (1) an \textit{exploration strategy} that determines which action to take at each time step given $c_t$ and some policy $\pi: c_t \rightarrow a_t$, and (2) a \textit{learner} that incorporates the bandit feedback received into the $\pi$.

\subsection{Learning: Online Linear Regression}

Assume that the true map $f^*$ exists in some function class $\mathcal{F}$. One method for solving the contextual bandit setting is to reduce the problem to regular online supervised learning and create an estimate of this function $\widehat{f}$ with least squares regression. Importance weighting \cite{Bietti2018} can eliminate the bias that comes from only using partial feedback from random exploration, but as our proposed exploration strategy is deterministic, we do not need to do this here.

Previous work \cite{2019Feng} demonstrated a model could accurately recommend the optimal action for a given food item in a full supervised learning setting. Subsequent work \cite{gordon2020adaptive} treated all but the last layer of SPANet as a fixed featurizer, treating the final linear layer as the context model. With this motivation, we assume that $f^*$ is linear and all observed noise is Gaussian, such that $l_t[a] = \theta_a^\top c_t + \epsilon$ with weights $\theta_a \in \R^{d_c}$ and noise $\epsilon \sim \mathcal{N}(0, \mathbf{I}\sigma^2)$. Applying least squares regression to this linear model allows us to produce familiar weight estimates
\begin{align}
    \widehat{\theta}_a = \left(\mathbf{C_a}^\top\mathbf{C_a}\right)^{-1}\mathbf{C_a}^\top L_a
\end{align}
where $\mathbf{C_a} \in \R^{T_a \times d_c}$ is the matrix contexts observed during the $T_a$ time steps where the agent selected action $a$, $L_a \in \R^{T_a}$ is the vector of scalar losses observed on those same time steps.

\subsection{Exploration: LinUCB}
\label{sec:linucb}
As described in Section \ref{sec:conban}, there are many existing exploration strategies for contextual bandit settings. We initially focus on the Linear Upper Confidence Bound (LinUCB) \cite{Chu2011} algorithm in the work since previous work \cite{gordon2020adaptive} suggests that it performs well empirically in the bite acquisition setting.

LinUCB implicitly balances exploration and exploitation by using the estimated linear model to construct a confidence interval around $l_t$ for a given $c_t$ and optimistically playing $a_t$ with the lowest lower confidence bound on its expected loss. In this way, the algorithm prefers relatively unknown actions with larger intervals (encouraging exploration) and actions with low loss (encouraging exploitation). UCB-style algorithms like this are known to achieve cumulative regret bounded by $\E[R_T] \leq \Tilde{O}(d_c\sqrt{T})$ \cite{Abbasi2011}.

For a given confidence level, this lower bound can be calculated as \cite{Chu2011}
\begin{align}
\label{eq:lcb}
    LCB(a) = \widehat{\theta_a}^\top c_t - \alpha\sqrt{c_t^\top\mathbf{\Sigma_a} c_t}
\end{align}
for some constant $\alpha > 0$ and $\mathbf{\Sigma_a} := \left(\mathbf{C_a}^\top\mathbf{C_a}\right)^{-1}$ the covariance of the estimator $\widehat{\theta_a}$.

We are further motivated to use LinUCB because, through this covariance matrix, it uses information about the learning scheme to tune its exploration. This is in contrast with learner-agnostic strategies like $\epsilon$-greedy. Therefore, when we modify the learning scheme with post hoc context, LinUCB will seamlessly incorporate the extra knowledge into its exploration strategy.
\begin{figure*}[t!]
    \centering
    \includegraphics[width=0.32\linewidth]{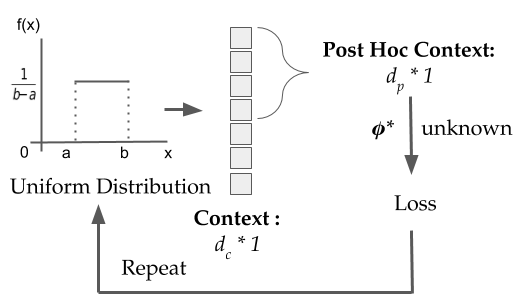}
    \includegraphics[width=0.35\linewidth]{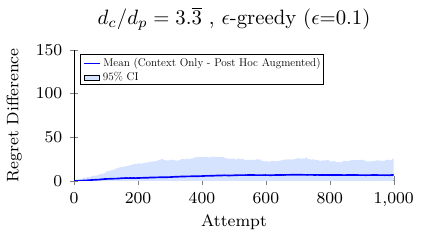} \hspace{-1.8em}
    \includegraphics[width=0.33\linewidth]{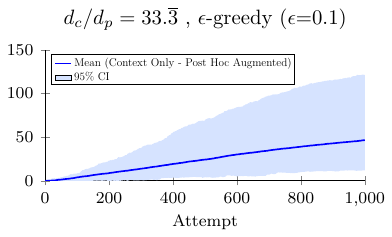}
    \vspace{-0.2cm}
    \caption{Experiment 1 shows the effect of dimensionality on the regret difference between the \textit{Context Only} and \textit{Post Hoc Augmented} model. The context vector is sampled from a uniform distribution and the post hoc vector is constructed using first $d_p$ components. The full loss vector is defined as $l_t = p_t^\top\phi^*$ where $\phi^*$ is unknown to the model. We observe an improvement in regret difference as $d_c$ is increased from 10 (\textit{Center}) to 100 (\textit{Right}).}
    \label{fig:synthetic}
    \vspace{-0.5cm}
\end{figure*}

\section{LEVERAGING POST HOC CONTEXT}
In general, we propose augmenting the conventional contextual bandit setting with an action-independent $d_p$-dimension post hoc context vector $p_t$ (haptic force-torque feedback in our setting) observable after selecting the action $a_t$. We can justify this inclusion as follows: At each time step, assume there exists some hidden state $z_t$ (for example, all the information describing an unripe banana slice). In our setting, the context $c_t$ (e.g., the picture of the food) and the post hoc context $p_t$ (e.g., haptic parameters only available after the action is taken) are just alternate representations of this underlying state (e.g., we can name the food item by either looking at it or touching it). Therefore, if there exists some function $h$ that maps the state onto the loss vector, then there should also exist functions $f$ and $g$ that map the context and post hoc context respectively onto that same loss vector. In other words, we assume $\E[l_t] = h(z_t) = f(c_t) = g(p_t)$.

This augmentation maps neatly onto the bite acquisition setting. At each round $t=1, \dots, T$, the interaction protocol consists of
\begin{compactenum}
	\item{\em Context observation.} The user selects a food item to acquire. We observe an RGBD image containing the single discrete food item (detected using RetinaNet \cite{2017Lin}). We pass this through SPANet \cite{gordon2020adaptive} which returns the visual context $c_t \in \R^{2048}$.
	\item{\em Action selection.} The algorithm selects a discrete manipulation strategy $a_t \in \mathcal{A} = \{1, 2, \dotsc, K\}$. In our initial implementation, $K=6$, matching a subset of the taxonomy of manipulation strategies from \cite{bhattacharjee2018food}. Action execution is detailed in Section \ref{sec:onrobot}.
	\item{\em Bandit loss observation.} The environment provides a binary loss $l_t(c_t)[a_t] \in \{0, 1\}$, where $l_t = 0$ corresponds to the robot successfully acquiring the single desired food item.
	\item{\em Post hoc context observation.} During action execution, time series force and torque data is passed through HapticNet (described below) to create the haptic context $p_t \in \R^{4}$.
\end{compactenum}

To expand on (4), HapticNet is a small multi-layer perceptron (MLP) from \cite{2019Gallenberger}. The first 50ms of force and torque data after contact with the food (as determined by force thresholding) are passed through two ReLu layers. The output is the softmax-ed vector $p_t$ classifying the food as ``hard'', ``medium'', ``soft'', or ``hard-skin''. Importantly, the hardness of the food is intrinsic, independent of the action. In \cite{2019Gallenberger}, the manipulation action used by human participants was directly affected by this categorization, and so we are motivated to use a linear model here as well.

The structure of this environment as it pertains to bite acquisition is shown in \figref{fig:system}.

\subsection{Learning: Joint Model Regression}
Recalling our assumption $\E[l_t] = f(c_t) = g(p_t)$, we propose jointly estimating $f$ and $g$ with least squares regression under the constraint that they produce the same outputs, i.e. $f(c_t)[a] = g(p_t)[a]\ \forall\ a$. This constraint could be a soft constraint, weighting the square difference between the outputs by hyperparameters. However, since the contextual bandit setting in general does not come with a well-defined training and validation set, it is desirable to reduce the number of hyperparameters requiring tuning. Therefore, in this work, we only consider using a hard constraint. Importantly, this constraint should be valid for all actions, allowing all time steps to factor into the estimate no matter which action was taken.

To demonstrate this, we jump into the linear setting, where $f(c_t)[a] = \theta_a^\top c_t$ and $g(p_t)[a] = \phi_a^\top p_t$.
\begin{align}
    \widehat{\theta}_a,\widehat{\phi}_a :&= \arg\min_{\theta_a, \phi_a}\twonormsq{\mathbf{C_a}\theta_a - L_a} + \twonormsq{\mathbf{P_a}\phi_a - L_a} \\ \text{s.t.}\ \mathbf{C}\theta_a &=  \mathbf{P}\phi_a
    \label{eq:hc}
\end{align}
As before, $\mathbf{C_a} \in \R^{T_a \times d_c}$, $\mathbf{P_a} \in \R^{T_a \times d_p}$, and $L_a \in \R^{T_a}$ are matrices of importance weighted contexts, post hoc contexts, and losses respectively. The constraint, being valid for all actions, uses the full context matrix $\mathbf{C} \in \R^{T \times d_c}$ and post hoc context data matrix $\mathbf{P} \in \R^{T \times d_p}$. Using the constraint to define a transformation matrix $\phi_a = (\mathbf{P}^\top\mathbf{P})^{-1}\mathbf{P}^\top\mathbf{C}\theta_a := \mathbf{H}\theta_a$, we can solve for the weight estimate.
\begin{align}
\label{eq:weights}
    \widehat{\theta}_a = \left[\mathbf{C_a}^\top\mathbf{C_a} + \mathbf{H}^\top\mathbf{P_a}^\top\mathbf{P_a}\mathbf{H}\right]^{-1}\left[\mathbf{C_a}^\top + \mathbf{H}^\top\mathbf{P_a}^\top\right]L_a
\end{align}

This formulation provides a normative reason to expect empirical improvements over the context-only setting. Consider the case exemplified by \figref{fig:intro}. If the post hoc context model is known perfectly, it can recommend the correct action for a given context after only a single attempt, cutting down exploration by a factor of $K$. More formally, if we know $\phi^*_a\ \forall\ a$, then we know what the expected loss $\E[L] = \mathbf{P}\phi^*_a$ would have been for action $a$ at all time steps, \emph{including time steps where we did not take action $a$}. In other words, we can rewrite the hard constraint from Equation \ref{eq:hc} as
\begin{align}
    \theta_a = (\mathbf{C}^\top\mathbf{C})^{-1}\mathbf{C}^\top\mathbf{P}\phi_a = (\mathbf{C}^\top\mathbf{C})^{-1}\mathbf{C}^\top \E[L]
\end{align}
This surface defined by the hard constraint is just the solution to standard linear regression on all time steps. The upshot is that knowing the post hoc context model reduces the problem from bandit feedback to full feedback regression, which is a much easier problem.

\subsection{Exploration: Modified LinUCB}
Under the linearity assumption, i.e., 
\begin{align}
    L_a = \mathbf{C_a}\theta_a^* + \epsilon = \mathbf{P_a}\phi_a^* + \epsilon
\end{align}
We can show that $\widehat{\theta}_a$ is an unbiased estimate of $\theta_a^*$ with a covariance bounded from above (via Cauchy-Schwartz and Jensen's inequalities) by
\begin{align}
\label{eq:sigma}
    \mathbf{\Sigma_p} := 2\left(\mathbf{C_a}^\top\mathbf{C_a} + \mathbf{H}^\top\mathbf{P_a}^\top\mathbf{P_a}\mathbf{H}\right)^{-1} 
\end{align}
From here, following the same logic as \cite{Chu2011}, we can construct a lower confidence bound equivalent to Equation \ref{eq:lcb} and replacing $\mathbf{\Sigma_a}$ with $\mathbf{\Sigma_p}$. This work focuses on comparing this modified algorithm to the baseline version of LinUCB as described in Section \ref{sec:linucb}.

Further details about the environment and algorithms in this section can be found at \cite{supplemental}. 
\begin{figure*}[t!]
    \centering
    \includegraphics[width=0.31\linewidth]{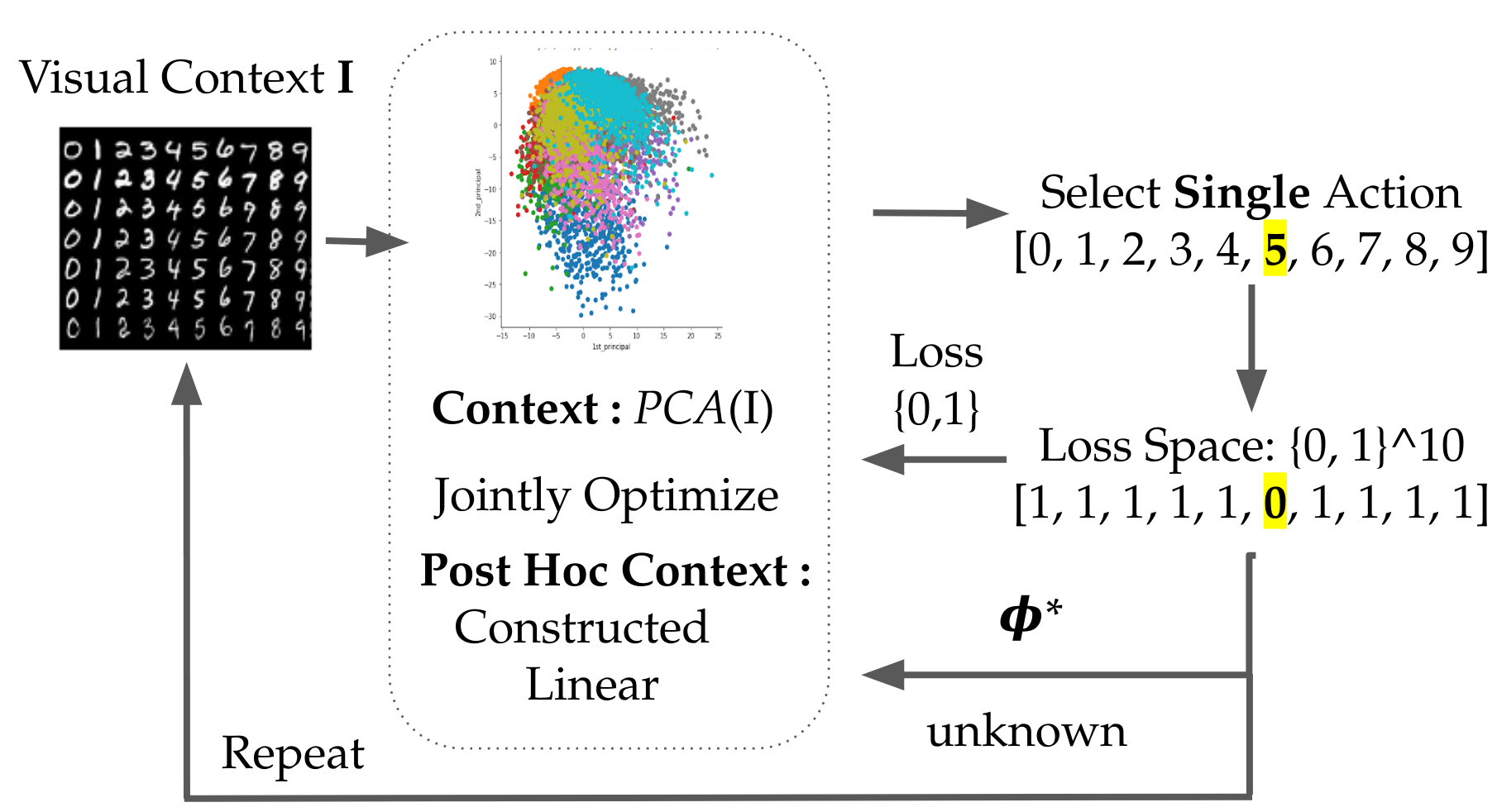}
    \includegraphics[width=0.32\linewidth]{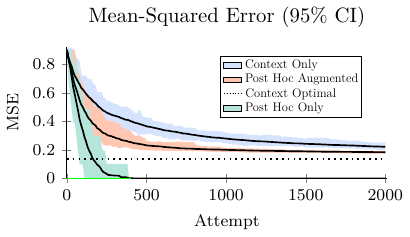}
    \includegraphics[width=0.34\linewidth]{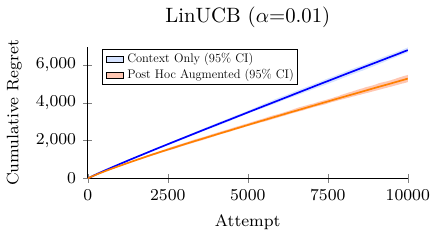}
    \caption{Experiment 2 is performed on the standard MNIST dataset to test learning speeds of various models. The context vector is reduced using the PCA and the post hoc context vector is constructed from the loss using $\phi^*$ which should be easy to learn. We see from the MSE plot (\textit{Center}) that the \textit{Post Hoc Augmented} model reaches it's optimal value much faster than the \textit{Context Only} model. As soon as the post hoc model is learned the context model quickly approaches it's best possible value. Also from the Regret plots (\textit{Right}) we can see the \textit{Post Hoc Augmented} model achieving lower regret value than it's \textit{Context Only} counterpart.}
    \label{fig:mnist}
    \vspace{-0.6cm}
\end{figure*}

\section{SYNTHETIC DATA EXPERIMENTS}
Before implementing this framework on the robot, we conducted two experiments with synthetic post hoc context to validate the potential benefits from this setting. Experiment 1 is designed to demonstrate that a low-dimension post hoc context vector can lead to faster learning, even if it contains no new information. It does so by varying the size of the synthetic context vector while keeping the size of the synthetic post hoc context vector fixed. Experiment 2 is designed to see if a well trained post hoc context model can effectively reduce the contextual bandit to the easier problem of full feedback online learning. The synthetic post hoc context is constructed to be easy to learn, while the context vectors, derived from MNIST data, do not even adhere to the linear model assumption. 
\subsection{Experiment 1: Low Dimension Post Hoc Context}
The setup for this experiment is outlined in \figref{fig:synthetic}(Left). We first fix the number of actions $K=10$ and the dimension of the post hoc context $d_p = 3$. We then generate a random, hidden pseudo-invertible post hoc context model $\phi^* \in \R^{d_p \times K}$. At each time step, we sample a context vector $c_t \sim [0,1)^{d_c}$, which is shown to the bandit algorithm. The first $d_p$ components are defined to be the post hoc context, and the full loss vector $l_t = p_t^\top\phi^*$ is computed accordingly. The algorithm incurs regret $l_t[a_t] - \min_al_t[a]$ and is shown $l_t[a_t]$ and $p_t$.

\paragraph*{Results}
For $d_c = 10$ and $d_c = 100$, we run the context-only and the post-hoc-augmented learners for 40 trials, 1000 attempts per trial, and record the cumulative regret. The results are shown in \figref{fig:synthetic}(Center, Right). With the lower dimensional context, we observe that the two learners perform comparably, with a slight advantage to the post-hoc-augmented learner. However, with the higher dimensional context, the post-hoc-augmented learner exhibits significantly better performance than the context-only learner. 

The context and the post hoc context contain the exact same information about the loss vector, but the post hoc context does so with fewer dimensions, making each observation more informative. These results support the idea that lower dimensional post hoc context is beneficial. Even if it contains no new information, it can still be used to train an accurate context model more quickly.


\subsection{Experiment 2: Easy-To-Learn Post Hoc Context}
\label{sec:exp2}
The setup for this experiment is outlined in \figref{fig:mnist}(Left). Context vectors are derived from the MNIST dataset \cite{lecun2010mnist}, which consists of labeled $28\times28$ images of hand-written digits (0 to 9) split into a training set with 60k samples and a test set with 10k samples. At each time step, we sample an image, and then use a PCA (trained on the training set) to reduce it to a $d_c = 200$ dimension context vector. This vector is shown to the bandit algorithm, which returns an action $a_t \in [K=10]$. For a given image, we can define the full loss vector $l_t \in \{0,1\}^{10}$ to be 1 if the incorrect digit is guessed and 0 otherwise. As in Experiment 1, we construct a random linear post hoc context model $\phi^* \in \R^{d_p \times K}$ and use it to manually construct a post hoc context vector $p_t \in \R^{d_p = 10}$ from the full loss vector. This makes for a post hoc context model that is extremely easy to learn perfectly. The loss $l_t[a_t]$ and the post hoc context $p_t$ is shown to the bandit algorithm.

\paragraph*{Results} First, to test learning speed, we ran a post-hoc-augmented learner and a context-only learner on the training set with random uniform exploration. Every 10 attempts, we freeze both linear context models $\widehat{\theta}$ and record its mean square error (MSE) on 2k samples from the test set.
\begin{align}
    MSE := \frac{1}{|\text{Test Subset}|}\sum_{c_t,l_t \in \text{Test Subset}}\twonormsq{\widehat{\theta}^\top c_t - l_t}
\end{align}
For the post-hoc-augmented learner, we also recorded the MSE of the post hoc context model $\widehat{\phi}$. All three are plotted in \figref{fig:mnist}(Center). Note that the best possible context model MSE (as determined by training on the full loss vectors from the entire training set) is 0.1383. We should expect no context model $\theta$ to beat that, even one augmented by post hoc context.

As expected, the noise-less post hoc model $\phi$ is learned perfectly as soon as it sees $d_p=10$ linearly independent samples for each digit (which happens within $\sim 500$ total samples). At this point, the post-hoc-augmented learner significantly deviates from the context-only learner, and it has reached its plateau within another $\sim 500$ samples. Meanwhile, the context-only learner still did not reach its plateau after 2000 samples. These results support the idea that a perfect context model effectively reduces the problem from bandit feedback to full feedback, allowing for faster learning.

We then combined each learner with a LinUCB exploration strategy, ran them on the entire 10k-sample test set, and recorded the cumulative regret. These results are shown in \figref{fig:mnist}(Right), and demonstrate a significant improvement of the post-hoc-augmented bandit over the context-only bandit. This shows that an improved learning speed can translate to reduced regret, the metric that we care about in this setting.
\begin{figure*}[t!]
    \centering
    \includegraphics[width=0.34\linewidth]{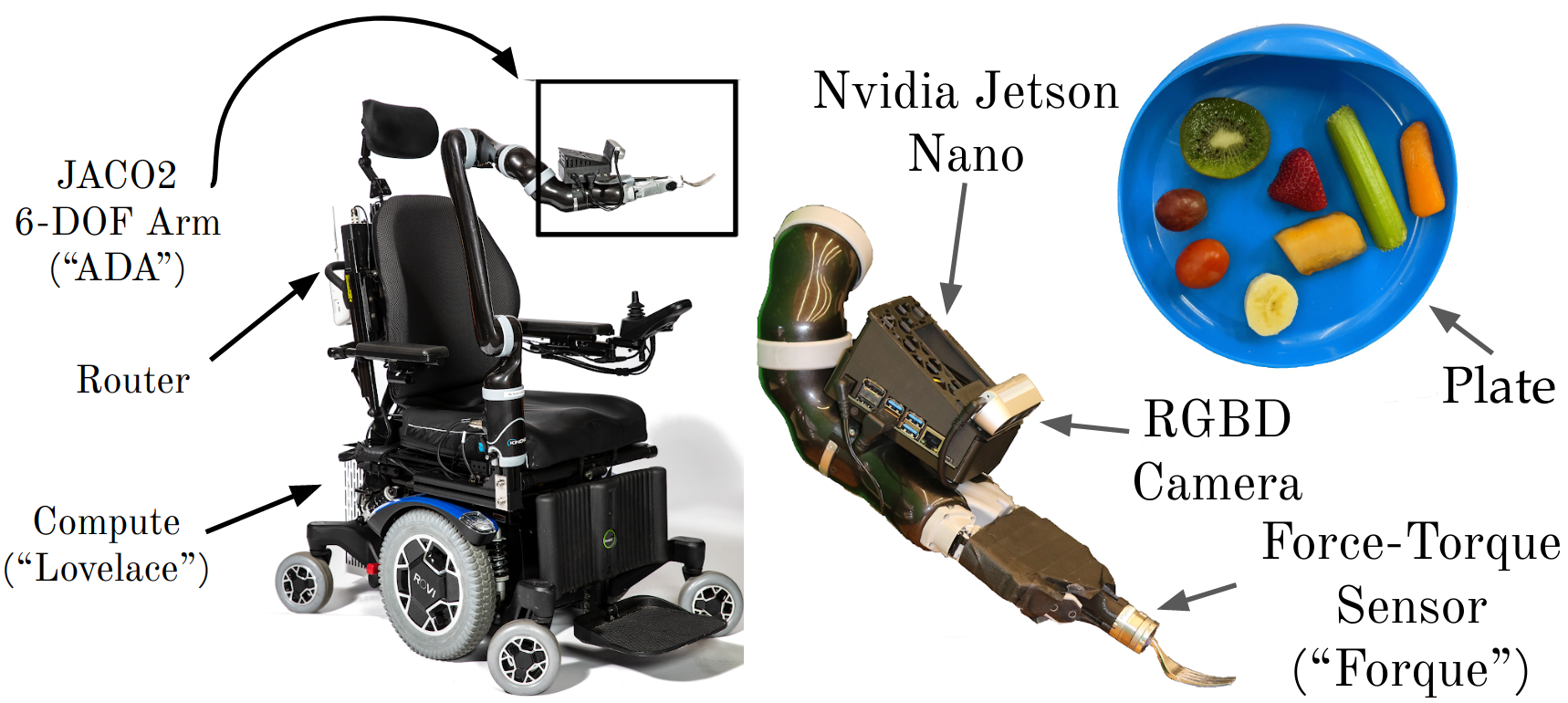}
    \includegraphics[width=0.31\linewidth]{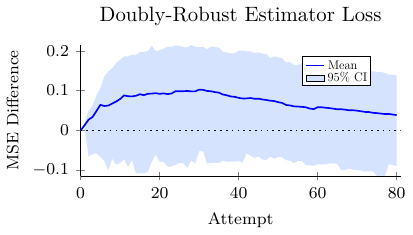}
    \includegraphics[width=0.30\linewidth]{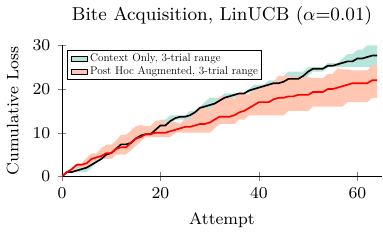}
    \vspace{-0.3cm}
    \caption{Results of the bite acquisition experiment using the Autonomous Dexterous Arm (ADA) (\textit{Left}). The MSE Difference plot (\textit{Center}) shows that there is an early benefit in learning with the post hoc context which reduces over time. Also from the Cumulative Loss plot (\textit{Right}) it's evident that with \textit{Post hoc Context Augmented} the cumulative loss incurred with increasing attempts is lower than it's \textit{Context Only} counterpart.}
    \label{fig:ada}
    \vspace{-0.6cm}
\end{figure*}
\section{REAL WORLD BITE ACQUISITION}
\subsection{System Description}
\label{sec:onrobot}
Our setup, the Autonomous Dexterous Arm (ADA) (\figref{fig:ada}(Left)), consists of a 6 DoF JACO2 robotic arm \cite{Jaco2018}. The arm has 2 under-actuated fingers that grab a custom-built, 3D-printed fork holder. For haptic input, we instrumented the fork with a 6-axis ATI Nano25 Force-Torque sensor \cite{Forque2018}. For visual input, we mounted a custom built wireless perception unit on the robot’s wrist; the unit includes the Intel RealSense D415 RGBD camera and the NVidia Jetson Nano for wireless transmission. Food is placed on a plate mounted on an anti-slip mat commonly found in assisted living facilities.

Each manipulation strategy is executed by a simple impedance controller with 3 parameters: (1) the angle of the fork handle relative to vertical, (2) the roll angle of the fork relative to the the major axis of the food item, and (3) the maximum force to impart on the food during skewering. Specifically, the 6 discrete actions in this setting, motivated by \cite{bhattacharjee2018food} and used in \cite{gordon2020adaptive}, are parameterized as follows
\begin{compactenum}
    \item \emph{Vertical Skewer 0 (VS0)}: Pitch 0, Roll 0, 25N
    \item \emph{Vertical Skewer 90 (VS90)}: Pitch 0, Roll $\frac{\pi}{2}$, 25N
    \item \emph{Tines Vertical 0 (TV0)}: Pitch -0.5, Roll 0, 20N
    \item \emph{Tines Vertical 90 (TV90)}: Pitch -0.5, Roll $\frac{\pi}{2}$, 20N
    \item \emph{Tilted Angled 0 (TA0)}: Pitch 0.4, Roll 0, 10N
    \item \emph{Tilted Angled 90 (TA90)}: Pitch 0.4, Roll $\frac{\pi}{2}$, 10N
\end{compactenum}

\subsection{Offline Results and Tuning}
The first step was to ensure that our post hoc context, the haptic data, was descriptive enough to potentially benefit the visual context model. To this end, we collected 115 samples of the robot skewering 3 food items, chosen to be representative of different haptic categories and optimal action (as determined in \cite{2019Feng}): grape is classified as ``hard skin'' and has the optimal action TV90, strawberry is ``medium'' and prefers VS0 or TV0, and banana is  ``soft'' and prefers TA0 or TA90. For each sample, we recorded the visual context $c_t$, post hoc context $p_t$, action taken $a_t$, loss $l_t[a_t]$, and food type name (e.g. ``grape'').

Since this data, by necessity, was collected with bandit feedback, we impute the full loss vector $\hat{l}$ of each attempt by averaging the loss for a given action across all samples of the same food type (e.g., average the loss of \emph{VS0} across all bananas). While simple, this can introduce a herding bias relative to the real world. We can eliminate this bias (at the cost of increased variance) by using a doubly robust \cite{Dudik2011} estimator
\begin{align}
\hat{l}_{DR}[a] = \hat{l}[a] + (l[i] - \hat{l}[a]) \frac{\1(i = a)}{\P[i = a]}
\end{align}
where $\hat{l}[a]$ is the biased estimate from herding, $\P[i = a]$ is the probability that we took action $a$ during data collection ($\frac{1}{6}$ in our case), and $l[i]$ is the actual binary loss associated with that sample (if available).

Similarly to Experiment 2 (Section \ref{sec:exp2}), we divided this data set into 80 training examples and 35 test examples. We then ran a context-only learner and a post-hoc-augmented learner on the training set with uniform exploration, freezing the context model $\widehat{\theta}_a$ after each time step to measure its MSE on the test set. The difference in MSE is shown in \figref{fig:ada}(Center). While the size and variance of the dataset makes it hard to show significance, the mean suggests that there may be a learning benefit to the post hoc context early on that fades over time.

We also used the full 115 samples to tune the exploration hyperparameter $\alpha$ for both LinUCB implementations. The optimal value for both the context-only and the post-hoc-augmented learners was $\alpha=0.01$, which we used for the online experiment.

\subsection{Online Experiment}
For the online experiment, we ran both the post-hoc-augmented LinUCB and the context-only LinUCB algorithms on 8 previously unseen food types, 8 attempts per food type, for a total of 64 attempts per trial. The 8 food types included 2 ``hard'' foods (carrot and celery), 2 ``medium'' foods (strawberry and cantaloupe), 2 ``soft'' foods (banana and kiwi), and 2 ``hard skin'' foods (cherry tomato and grape). In the absence of noise, we would expect the robot to take about 48 attempts to figure out the optimal action for each food item: 6 actions for each of the 8 food items. Given that we expect the optimal action to be knowable from the haptic category, we would hope to decrease that convergence time by a factor of 2 with the post hoc context: 6 actions for each of the 4 haptic categories.

The results of the experiment are shown in \figref{fig:ada}(Right). In this setting, where it is impossible to observe the full loss vector $l_t$, we record cumulative loss $\sum_tl_t[a_t]$ instead of cumulative regret. By this metric, we do see some improvement of the post-hoc-augmented agent over the context-only agent. Over 3 trials, the former experienced fewer failures, accumulating an average total loss of 22 compared with 27.667 for the context-only learner.
\section{DISCUSSION}
One key takeaway from this work is the success of post-hoc-augmented bandits in a variety of empirical settings that potentially deviate significantly from the assumed linear model.  This suggests that it may be fruitful to pair this augmentation with other empirically competitive exploration strategies such as Thompson Sampling.

With that said, a limitation of this work is the requirement that the post hoc context be independent of the action selected. While this happens to work in robot-world physical interactions like bite acquisition, it is harder to imagine it working in a domain like online recommendation systems, where a user's experience (and thus, the feedback they can give as post hoc context) will be affected by, for example, the set of links they were shown. Future work should address this setting and potentially relax our strong assumption that $\E[l_t] = f(c_t) = g(p_t)\ ;\ \forall a$.

Finally, we note that our current action space is small and imperfect. We cannot expect to converge to 0-loss on all possible food items that a user might want to eat. In the future, we intend to broaden our scope to larger classes of food items by investigating continuous and expanded action and post hoc context spaces rather than limiting ourselves to expert-defined discrete options. It is unclear how to apply post hoc context to this environment. One possibility is to treat it as a function of the gradient of the loss function with respect to the action space $\nabla_a l_t$. We could also expand the action space by considering compound (or slate \cite{Dimakopoulou2019}) actions. The space of possible food items and acquisition strategies is massive, and users require a robust system if they need to use it daily \cite{bhattacharjee2020userpref}, so a lot more research can be done here.

Overall, these results suggest that multi-modal feedback can be leveraged in interactive learning to allow for more data-efficient adaptive bite acquisition.

\section*{ACKNOWLEDGMENTS}
Research reported in this publication was supported by the Eunice Kennedy Shriver National Institute Of Child Health \& Human Development of the National Institutes of Health under Award Number F32HD101192. The content is solely the responsibility of the authors and does not necessarily represent the official views of the National Institutes of Health. This work was also (partially) funded by the National Science Foundation IIS (\#2007011), National Science Foundation DMS (\#1839371), the Office of Naval Research, US Army Research Laboratory CCDC, Amazon, and Honda Research Institute USA.

\bibliographystyle{IEEEtran}
\bibliography{IEEEabrv,main}
\end{document}